\def\year{2022}\relax
\documentclass[letterpaper]{article} % DO NOT CHANGE THIS
\usepackage{aaai22}  % DO NOT CHANGE THIS
\usepackage{times}  % DO NOT CHANGE THIS
\usepackage{helvet}  % DO NOT CHANGE THIS
\usepackage{courier}  % DO NOT CHANGE THIS
\usepackage[hyphens]{url}  % DO NOT CHANGE THIS
\usepackage{graphicx} % DO NOT CHANGE THIS
\usepackage{natbib}  % DO NOT CHANGE THIS AND DO NOT ADD ANY OPTIONS TO IT
\usepackage{caption} % DO NOT CHANGE THIS AND DO NOT ADD ANY OPTIONS TO IT
\usepackage{amsmath}
\usepackage{float}
\usepackage{subcaption} 

\DeclareCaptionStyle{ruled}{labelfont=normalfont,labelsep=colon,strut=off} % DO NOT CHANGE THIS
\usepackage{algorithm}
\usepackage{algorithmic}

\title{Toward Semantic Scene Understanding for Fine-Grained 3D Modeling of Plants}
\author{
    Mohamad Qadri,
    Harry Freeman, 
    Eric Schneider, 
    George Kantor 
}
\affiliations{
    Carnegie Mellon University, Robotics Institute\\
    5000 Forbes Ave, Pittsburgh, PA 15213
}

\usepackage{bibentry}
\begin{document}

\maketitle

\begin{abstract}
% Visual Simultaneous Localization and Mapping (SLAM) and accurate 3D mapping systems are essential components in agricultural robotics that allow for autonomous navigation and accurate plant health and yield estimates.  However, lack of texture, varying illumination conditions, and lack of structure in the environment pose a challenge for visual SLAM and 3D reconstruction systems that rely on traditional feature extraction and matching algorithms such as ORB and SIFT.  This paper proposes that using semantic data in mapping and reconstruction improves results compared to traditional methods. Specifically, we 1) use sorghum seeds as semantic landmarks to build a SLAM based system that enables us to map 78\% of a sorghum range on average, compared to 38\% with ORB-SLAM2, and 2) use seeds as semantic features to improve 3D reconstruction of a full sorghum panicle.

Agricultural robotics is an active research area due to global population growth and expectations of food and labor shortages.
Robots can potentially help with tasks such as pruning, harvesting, phenotyping, and plant modeling. However, agricultural automation is hampered by the difficulty in creating high resolution 3D semantic maps in the field that would allow for safe manipulation and navigation.
% \item Our end goal is scene-aware XXXXXXX and manipulation of plants, which rests on a foundation of semantic scene understanding and reasoning
In this paper, we build toward solutions for this issue and showcase how the use of semantics and environmental priors can help in constructing accurate 3D maps for the target application of sorghum.
Specifically, we 1) use sorghum seeds as semantic landmarks to build a visual Simultaneous Localization and Mapping (SLAM) system that enables us to map 78\% of a sorghum range on average, compared to 38\% with ORB-SLAM2; and 2) use seeds as semantic features to improve 3D reconstruction of a full sorghum panicle from images taken by a robotic in-hand camera.
\end{abstract}

\section{Introduction}

Imagine a fully automated mobile manipulator with two cooperative robotic arms tasked to create a full 3D reconstruction of all fruits in a tree canopy, with some fruit initially occluded by branches and leaves. One arm pushes a branch aside while the other arm moves through free space to take images of the exposed area. Our vision is to move towards developing such a system. The first step towards this goal is to develop algorithms that can understand and reason about 3D semantics in the scene to allow for safe and reliable manipulation. This requires accurate high-resolution 3D reconstruction. 

Existing 3D reconstruction, visual Simultaneous Localization and Mapping (SLAM), and Structure from Motion (SFM) algorithms fundamentally rely on the accuracy of traditional visual feature matching methods, such as SIFT \cite{sift} and ORB \cite{orb} (used by popular feature-based SLAM methods such as ORB-SLAM2 and ORB-SLAM3). These features perform poorly in agricultural environments due to the lack of texture in the images, variations in luminosity levels, and the dynamics of the environment (for example, leaves or crops moving due to wind). 

In this paper, we demonstrate 1) how the use of semantics and environmental constraints, such as the structure of robotic navigation in agricultural fields, enables the development of robust SLAM systems for 3D mapping in agriculture and 2) how semantics can improve ICP-based registration for high-definition 3D modeling of plants. We focus on two target applications: mapping in sorghum fields and full panicle 3D reconstruction using a robotic arm with an in-hand stereo camera.

\section{Related Work}
%For manipulation, \cite{arm} applies priors towards an amodal 3D reconstruction system to improve manipulation in cluttered environments. 
%In visual-based SLAM, there have been significant progress in both direct \cite{lsd-slam} and indirect \cite{orbslam2} methods.  However, these methods fail to extend to agricultural settings. \cite{fusion++}, \cite{ok2019robust}, \cite{rols}, \cite{slam++}, and \cite{choudhary2014slam} used learned features and landmarks, sometimes with scene structure or prior model assumptions, but these approaches are either indoor or have less cluttered scenes. \cite{quadric} and \cite{cubic} use semantic SLAM methods but these perform poorly in cluttered environments. There are several impressive 2D object detection [\cite{yolov4}, \cite{ssd}, \cite{Faster-RCNN}] and semantic segmentation networks [\cite{pspnet}, \cite{Mask-rcnn}].  In agriculture, \cite{stalknet} and \cite{parharcgan} use segmentation to measure stalk width. \cite{rols} builds a SLAM system using geometric primitive shapes fitted to grapes as semantic 3D landmarks for SLAM. \cite{sem_plan_trav} uses semantic segmentation to determine plant traversability, and \cite{monofruit} uses semantic data for point cloud alignment to count apples. \cite{assigning_apples}, \cite{mola_fruit_detect_locate} and \cite{stein_yield} tackle similar issues for apples and mangoes with point cloud matching while \cite{SANTOS2020105247} uses semantic features for grape counting.

There has been significant progress in visual SLAM for both direct \cite{lsd-slam} and indirect \cite{orbslam2}, \cite{campos2021orb} methods.  However, these methods fail to extend to agricultural settings due to varying lighting conditions and repeated patterns. \cite{fusion++}, \cite{ok2019robust}, \cite{choudhary2014slam}, \cite{quadric} use learned features and landmarks, sometimes with scene structure or prior model assumptions, but these approaches are either indoor or assume less cluttered scenes.
% use semantic SLAM methods but these perform poorly in cluttered environments.
There are several impressive 2D object detection \cite{yolov4}, \cite{ssd}, \cite{Faster-RCNN} and semantic segmentation \cite{pspnet}, \cite{Mask-rcnn} networks which various works build upon. In agriculture, \cite{stalknet} and \cite{parharcgan} use segmentation to measure stalk width. 
%\cite{mccool2016visual} propose a novel method to segment highly occluded crops, applied to segmenting sweet peppers. 
\cite{rols} build a SLAM system using geometric primitive shapes fitted to grapes as semantic 3D landmarks. 
%\cite{sem_plan_trav} uses semantic segmentation to determine plant traversability, and 
\cite{monofruit} uses semantic data for point cloud alignment to count apples. \cite{dong2020semantic} and \cite{SANTOS2020105247} tackle similar issues for apples and grapes. \cite{sodhi2018robust} uses a robotic system to create in-field 3D reconstructions of sorghum plants. \cite{sepulveda2020robotic} present segmentation, planning, and occlusion algorithms to increase the picking accuracy of a dual-arm aubergine harvesting robot. \cite{zine2021assigning} presents a method to delineate apple trees in a trellis structured orchard
and perform fruit count. We build on top of these works and present promising results along with future research directions to promote in-field 3D semantic mapping and safe manipulation in agriculture.

\section{Semantic Features as SLAM Landmarks}
\label{sectionslam}
In this section, we demonstrate how using semantics, leveraging environment specific constraints, and reasoning about the geometry of the scene can alleviate some of the the 3D reconstruction and data association challenges in agriculture. 
One such prior is robotic navigation in agricultural environments; robots traverse the field one row at a time, generally moving in a straight line. We show how incorporating this prior and assumptions about the geometric relationship between semantic landmarks leads to improvements in data association accuracy and hence increased SLAM robustness. Our focus application is 3D mapping in sorghum fields using sorghum seeds as semantic landmarks. 

\subsection{Semantic SLAM Leveraging Robotic Navigation Constraints}
\label{semanitcslam}
% A diagram of the SLAM system is shown in Fig. \ref{fig:slam_pipeline} (appendix).
This section describes the front-end system where semantic and geometric constraints are enforced. The full SLAM system is further described in \cite{Qadri-2021-128985, qadri2021semantic}.
\label{slam_front_end}
\subsubsection{Feature Extraction}
%There have been significant advances in deep-learning based object detection and segmentation.  However, as a result of the lack of available datasets in agriculture, it is hard to train networks that generalize to varying conditions. One method we use to approach this issue using an illumination-invariant camera system from \cite{abhi_stereo} that allows training with less data by keeping image illumination consistent across different environmental conditions.

\begin{figure}[h!]
\centerline{\includegraphics[width=0.8\linewidth]{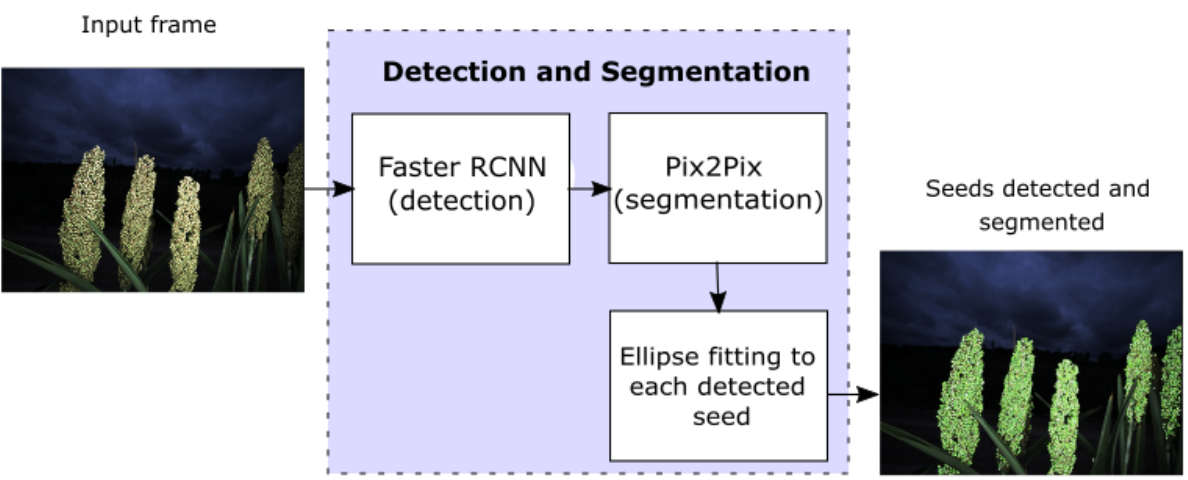}}
\caption{Detection and segmentation pipeline.}
\label{fig:detect_seg}
\end{figure}

The feature extraction pipeline (Fig. \ref{fig:detect_seg}) is based on \cite{parharcgan}. A Faster-RCNN network with a VGG16 backbone is used for detection, and returns a bounding box for each seed in the image. Each bounding box is cropped and passed to a pix2pix \cite{pix2pix} network, which generates a new image with a segmentation mask for each detected seed. 
%Pix2pix was chosen because it uses less training data than other methods like Mask RCNN.
After segmentation, a 2D ellipse is fitted to the segmented areas, and ellipse centers are used as semantic keypoints.

\subsubsection{Data association algorithm}
The object-level data association between stereo pairs and successive temporal frames is framed as linear sum assignment problem (LSAP) optimization \cite{assignment}. 
We define a bipartite graph $G = (U, V, E)$. Each vertex $s_{ab} \in U$, with coordinates $(a,b)$ in the camera frame, corresponds to the projection of a 3D landmark onto image A. Similarly, each vertex $s_{mn}$ with coordinates $(m,n) \in V$ corresponds to a projection  onto image B. An edge $c_{ij} \in E$ between nodes $s_{ab}$ and $s_{mn}$ defines the cost of associating $s_{ab}$ to $s_{mn}$. By introducing an assignment matrix $\varphi$ where $\varphi_{ij} \in \{0,1\}$, LSAP can be framed as the following optimization problem:

\begin{figure}[h!]
\centerline{\includegraphics[width=0.9\linewidth]{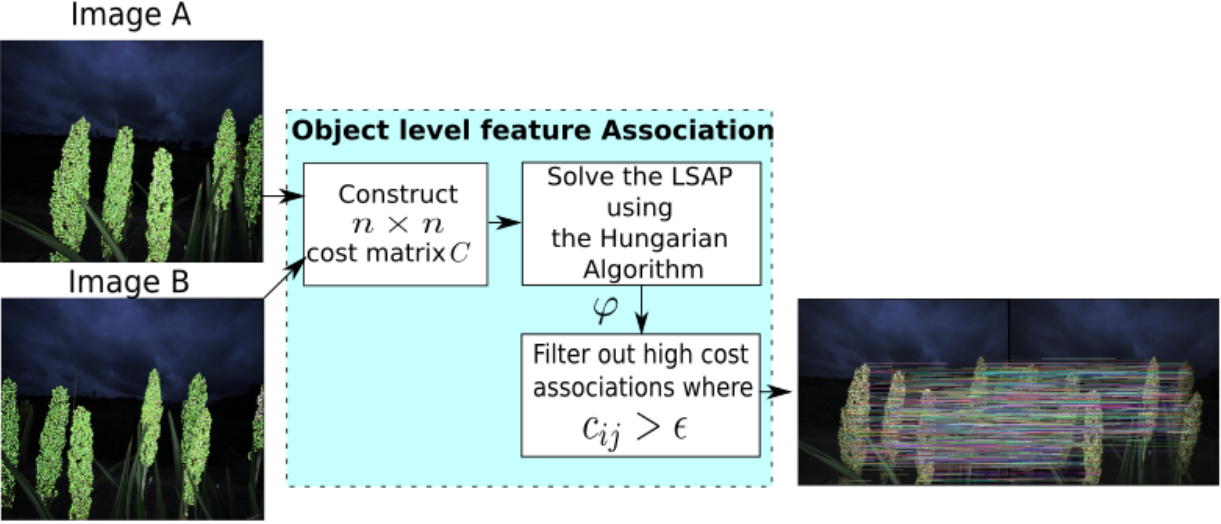}}
\caption{Proposed feature association pipeline.}
\label{fig:data_association}
\end{figure}

\begin{align*}
    \min_{\varphi \in S} \sum_{i=1}^N \sum_{j=1}^N c_{ij} \varphi_{ij} \
        \text{subject to}:    \sum_{j=1}^N \varphi_{ij} = 1 ,\ i \in U \\
     \sum_{i=1}^N \varphi_{ij} = 1 ,\ j \in V
\end{align*}
Where $S$ is the set of all possible assignments of nodes in $U$ to nodes in $V$. Since sorghum panicles are rigid bodies, the distance from a particular seed to its neighboring seeds should stay approximately constant as the robot moves in the environment. Hence, we add the constraint that the sum of the Euclidean distances of a node $s_{ab} \in U$ to its surrounding nodes in $U$ should be approximately equal to the sum of the Euclidean distances of $\varphi^*(s_{ab})$ $\in V$ and its surrounding nodes in $V$, where $\varphi^*$ is the optimal assignment. We define a heuristic cost function that captures this geometric structure between the landmarks. For each node $s_{ab} \in U $, we define sets of neighbouring nodes: $L_{ab}$ (left), $R_{ab}$ (right), $T_{ab}$ (top), and $B_{ab}$ (bottom) satisfying the conditions:
\begin{align*}
L_{ab} = \{\forall s'=(c,d) \in U \: | \: 0 < a - c < \Delta \: \text{and} \: |d-b| < \epsilon\} \\ 
R_{ab} = \{\forall s'=(c,d) \in U \: | \: 0 < c - a < \Delta \: \text{and} \: |d-b| < \epsilon\} \\ 
T_{ab} = \{\forall s'=(c,d) \in U \: | \: 0 < b-d < \Delta \: \text{and} \: |c-a| < \epsilon\} \\ 
B_{ab} = \{\forall s'=(c,d) \in U \: | \: 0 < d-b < \Delta \: \text{and} \: |c-a| < \epsilon\}
\end{align*}

We define $L_{mn}$, $R_{mn}$, $T_{mn}$, $B_{mn}$ similarly for $s_{mn} \in V $. The cost of associating node $s_{ab}$ to node $s_{mn}$ is defined as:
\begin{align*}
    C(s_{ab}, s_{mn}) = r \cdot C'(L_{ab}, L_{mn}) +  r \cdot C'(R_{ab}, R_{mn}) + \\
    r \cdot C'(B_{ab}, B_{mn})+ r \cdot C'(T_{ab},T_{mn}) + |b-n|  \\
  \text{where }C'(X, Y) = \frac{\sum_{s' \in X} \sqrt{(s'_x - a)^2 + (s'_y - b)^2}}{\sum_{s' \in Y} \sqrt{(s'_x - m)^2 + (s'_y - n)^2}} \\ 
\end{align*}
$r$ is a constant, and $|b - n|$ is a term added to penalize matching landmarks with high vertical difference due to the horizontal nature of the robot trajectory. $s'_x$ and $s'_y$ are the $x$ and $y$ coordinates of one of the surrounding nodes $s'$. 
%The Hungarian algorithm \cite{assignment} solves the LSAP in $O(n^3)$. Efficient CUDA-based implementations running on Nvidia GPUs  \cite{gpuhungarian2} provide significant speedups compared to CPU-based implementations.
 \subsubsection{Cost as a matching confidence measure}
The Hungarian algorithm returns the optimal assignment matrix $\varphi^*$, which is a bijection from $U$ to $V$. Each row in $\varphi^*$ is a one-hot vector, where $\varphi_{ij}^*=1$ indicates node $i$ has been matched with node $j$, and has a cost $c_{ij}$. Removing assignments where $c_{ij}$ is over a threshold keeps only high confidence matches.

%\subsection{SLAM Back-end} 

\subsection{SLAM Results}
\begin{figure}[h]
\centering
\includegraphics[width=0.7\columnwidth]{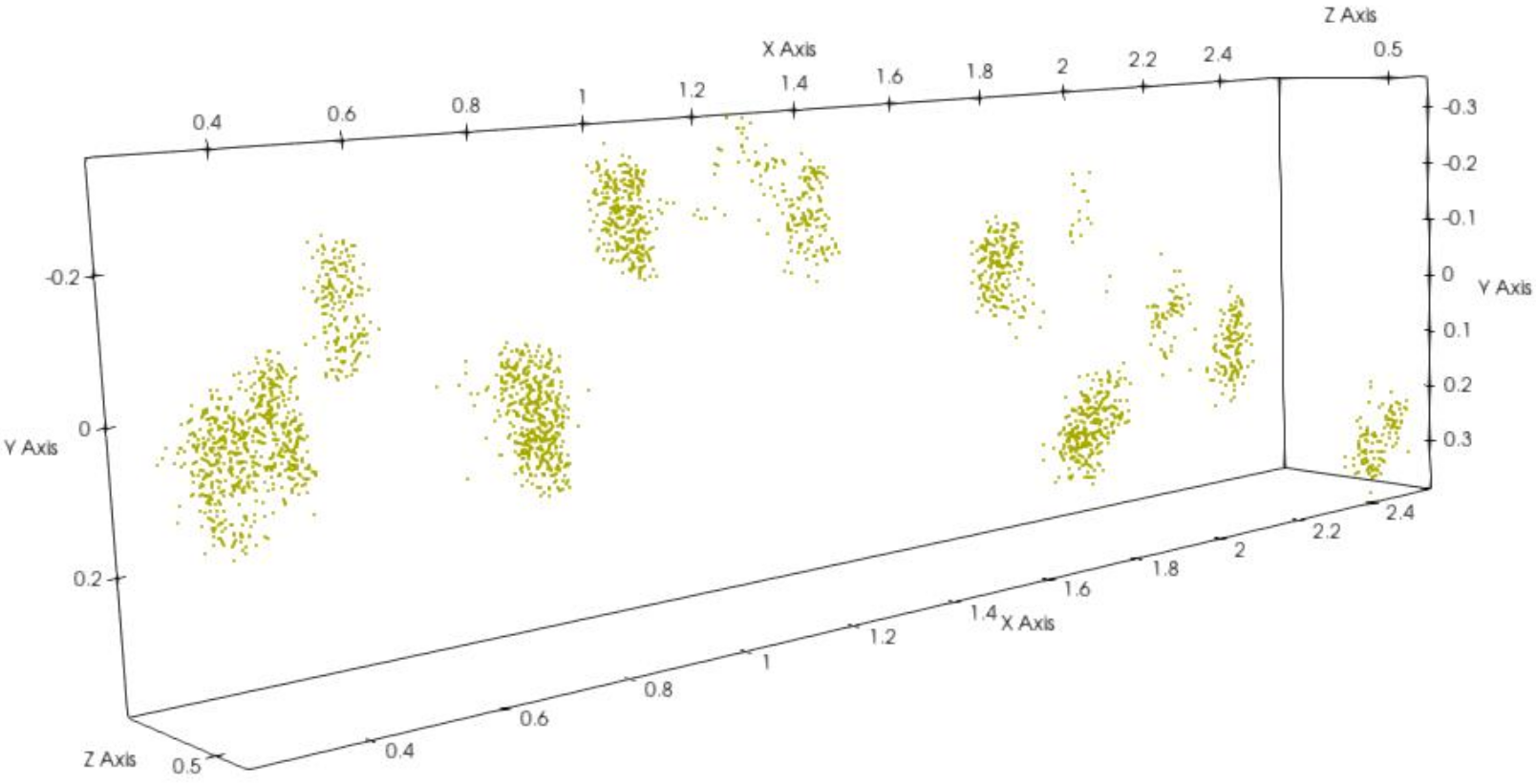} 
\caption{Example of a reconstructed 3D scene. The green dots correspond to 3D sorghum seeds and grouping of points correspond to different sorghum panicles. The first 1m
is reconstructed from the sequence of images in Fig. \ref{fig:series} (appendix).}
\label{fig1}
\end{figure}

Once features are detected and associated, we frame the back-end optimization as a standard factor-graph problem which returns the optimized 3D landmark locations and camera trajectory. Fig. \ref{fig1} is an example of a reconstructed sorghum range\footnote{Sorghum fields are composed of rows, each containing several ranges. A range is $\approx$4m long and may contain different varieties of sorghum.  Empty spaces with no plants separate consecutive  ranges.}.
% The first meter of the range is constructed from the sequence of images in Fig. \ref{fig:series} (appendix).
We use Maximum Distance Mapped as an indicator metric for the stability of SLAM systems and performance of data association algorithms. This is the distance that SLAM was able to map before a failure occurred in the back-end optimization or the system lost track. Table \ref{table:mdm} compares the maximum distance mapped with our semantic features against traditional descriptors with a brute force matcher (BF). Using our proposed matching algorithm, we can map 3 out of 8 sorghum ranges
completely and map $65\%$ of the remaining 5 ranges on average ($78\%$ on average across the 8 sorghum ranges). SIFT performed the best out of the remaining four algorithms, with which we are able to map around $38\%$ of the 8 sorghum ranges on average. This shows that using geometric relationships between semantic landmarks can improve performance when visual feature descriptors perform poorly. 

\begin{table}[h!]
    \begin{center}
    \resizebox{1\linewidth}{!}{
        \begin{tabular}{|c|c|c|c|c|c|}
            \hline
            \textbf{} & \multicolumn{5}{|c|}{\textbf{Feature detector + Matcher}} \\
            \cline{2-6} 
            \textbf{Range ID} \\ (length in m) & \textbf{\textit{SIFT + BF}}& \textbf{\textit{SURF + BF}}& \textbf{\textit{ORB + BF}} & \textbf{\textit{AKAZE + BF}} & \textbf{\textit{OURS}} \\
            \hline
            \textbf{1} (3.56 m)  & 1.86 m & 1.55 m &  0.2 m &  1.55 m &  \textbf{3.56 m} \\
            \hline
            \textbf{2} (5.00 m)  & 0.25 m & 0.25 m&  Failed &  0.19 m  &  \textbf{5.00 m} \\
            \hline
            \textbf{3} (4.42 m) & 0.5 m & 0.38 m & Failed &  0.38 m  &  \textbf{2.85 m}  \\
            \hline
            \textbf{4} (4.1 m)  & 1.47m & 0.6 m  &  Failed &  1.14 m &  \textbf{2.31 m} \\
            \hline
            \textbf{5} (4.78 m) & 0.57 m & 0.74 m &  failed &  0.74 m &  \textbf{2.31 m}  \\
            \hline
            \textbf{6} 3.94 m  & 1.4 m & 0.19 m &  0.11 m  &  0.45 m &  \textbf{3.2 m} \\
            \hline
            \textbf{7} 5.03 m & 3.15 m & 0.26 m &  Failed &  0.46 m &  \textbf{3.72 m}  \\
            \hline
            \textbf{8} 4.43 m & 4.04 m & 0.94 m &  Failed &  0.94 m &  \textbf{4.43 m} \\
            \hline
        \end{tabular}
    }
    \caption{Maximum distance mapped. For fair comparison, we remove all matches that do not adhere to the camera motion assumptions (horizontal travel) for all methods. All  ground truth distances are extracted from GPS. }%\textit{BF} indicates brute force matching.}
    \label{table:mdm}
    \end{center}
\end{table}
In Table \ref{table:orb-slam2-comp}, we compare the performance of our SLAM algorithm against ORB-SLAM2, a feature-based SLAM system using DBoW2 for feature matching. We report the maximum distance mapped before the system become ``lost".

%We expect other feature-based methods such as \cite{campos2021orb} to perform similarly which further motivates the use of semantic features.
\begin{table}[H]
\label{tab2}
\begin{center}
\resizebox{0.55\linewidth}{!}{%
\begin{tabular}{|c|c|c|}
\hline
\textbf{} & \textbf{\textit{ORB-SLAM2}}& \textbf{\textit{OURS}}\\
\hline
\textbf{Range 1} (3.56m) &  0.35m  &  \textbf{3.56m} \\
\textbf{Range 2} (5.00m) &  0.25m  &  \textbf{5m} \\
\textbf{Range 3} (4.42m) &  0.18m  &  \textbf{2.85m} \\
\textbf{Range 4} (4.10m) &  0.25m  &  \textbf{2.31m} \\
\textbf{Range 5} (4.78m) &  0.31m  &  \textbf{2.31m} \\
\textbf{Range 6} (3.94m) &  0.12m  &  \textbf{3.2m} \\
\textbf{Range 7} (5.03m) &  0.33m  &  \textbf{3.72m} \\
\textbf{Range 8} (4.43m) &  0.26m  &  \textbf{4.43m} \\
\hline
\end{tabular}
}
\caption{ORB-SLAM2 vs. OURS}
\label{table:orb-slam2-comp}
\end{center}
\end{table}

These results illustrate the expected performance of feature-based SLAM methods when feature descriptors, a fundamental building block, perform poorly due to  lack of texture and variations in luminosity levels, which further motivates the use of semantic features. We also ran ORB-SLAM3 on our dataset which improves on the re-localization capabilities of ORB-SLAM2 by building local maps when the system is lost. Local maps are merged when revisiting already mapped areas. We observed that ORB-SLAM3 performs similarly to ORB-SLAM2; the system repeatedly enters one of the ``lost" states every few frames, indicating that ORB-SLAM3 is only able to construct local maps using a few images before losing track again.

% Indeed, we ran ORB-SLAM3 \cite{campos2021orb} on our dataset, which adds better re-localization capabilities when the system loses track. We observed that the system enters one of the ``lost" states every few frames, indicating that ORB-SLAM3 is unable to construct a map using more than a few images.

% \subsection{SLAM as a first step}
% We have demonstrated how using additional problem specific constraints, in this case horizontal travel, assumptions about panicle rigidity, and the choice of seeds as a first-class feature, can lead to improved performance of SLAM system. Going forward, the system could be improved by considering additional relevant information in the feature matching step. For example, learned deep visual features such as <CITE PAPER(S) HERE> can be incorporated with application specific geometric constraints for a more robust data association process. 
%For example, there is a great deal of literature on 2D feature matching, where deep features are state of the art. Learned feature similarity could be incorporated into the seed matching cost function to combine the designed geometric constraints with visual features. However good mapping from a camera in the middle of a row gets though, there will still be occlusion issues when counting seeds due to the restricted panicle views. The next section discusses initial attempts to model a full panicle.

\section{High Resolution 3D Modeling with Semantic Features}
\begin{figure*}[h!]
\begin{subfigure}[h]{0.8\textwidth}
\includegraphics[width=1\columnwidth]{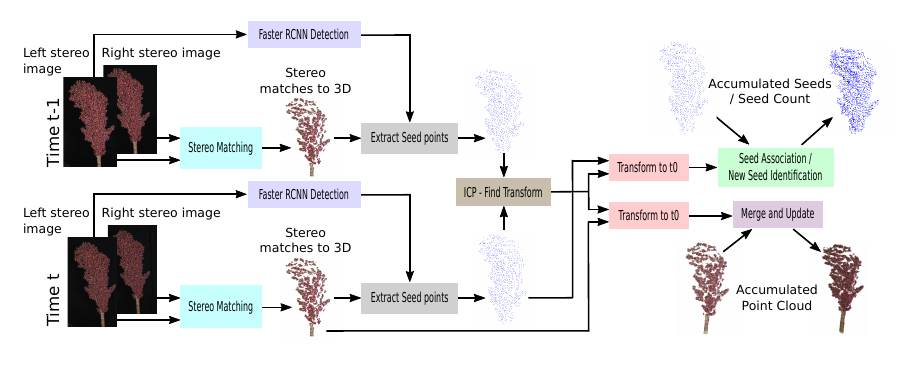}
\caption{3D Reconstruction and Seed Matching System}
\label{reconstruction_pipeline}
\end{subfigure}
\begin{subfigure}[h]{0.15\textwidth}
\includegraphics[width=1\columnwidth]{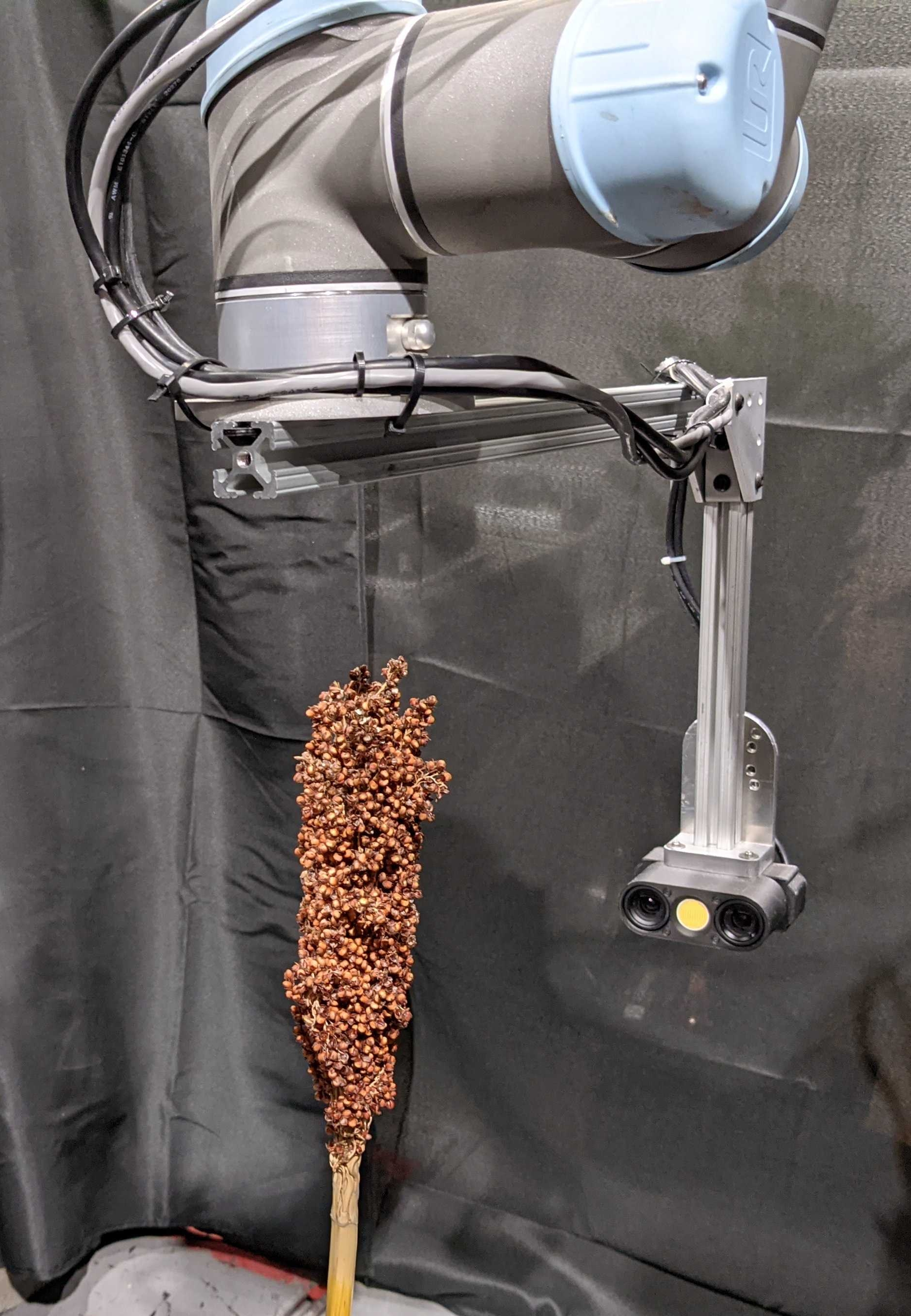}
\caption{Prototype end-effector for a UR5 allowing a $360^{\circ}$ view of the panicle. Camera from \cite{abhi_stereo}.}
\label{capture}
\end{subfigure}
\caption{}
\end{figure*}

% \section{Mobile Manipulation for Generating Complete 3D Reconstructions}
% \section{Proposed Reconstruction of Sorghum Panicles For Seed Counting}
In the previous section, we presented a system with a camera rigidly attached to a mobile robot moving in the environment, commonly used in agricultural robotics. However, a fundamental shortcoming of this approach is its inability to build full 3D reconstructions because it is limited to mapping the visible face of an object. This is a substantial limitation since occlusions are common in agricultural settings. Hence, development of novel methodologies on the system and algorithmic levels should be made to reason and deal with such occlusions. In this section, we propose a process in which a robotic arm with an attached in-hand camera (Fig. \ref{capture}) is able to capture a full $360^{\circ}$ set of stereo images of a single sorghum panicle and calculate phenotyping data. We qualitatively show how combining forward kinematics (FK) with semantic features can improve 3D reconstruction. We plan to eventually evaluate reconstruction and seed matching accuracy on a surrogate metric, seed count.

% \begin{figure}[h]
% \centering
% \includegraphics[width=0.3\columnwidth]{LaTeX/sorghum_scanning_system.jpg}
% \caption{Prototype end-effector for a UR5 allowing a $360^{\circ}$ view of the panicle. Camera from \cite{abhi_stereo}.}
% \label{fig:capture}
% \end{figure}

%However good mapping from a camera in the middle of a row gets, there will still be occlusion issues 
%when counting seeds due to the restricted panicle views. The next section discusses initial attempts to model a full panicle

%In pursuit of a full seed count for a given panicle, we are taking stereo images of all faces of the panicle, merging the views, and then using that 3D model to eliminate duplicated seed detections from 2D segmentation algorithms. For this approach to be workable the reconstruction needs to be extremely accurate, since seeds are generally 2-4 mm across and elimination of duplicates based on 3D position requires placement roughly within this size limit. Our best reconstruction results so far have come from using segmented seeds as the base features for Iterative Closest Point (ICP).

% \begin{figure*}[ht!]
% \centering
% \includegraphics[width=0.75\linewidth]{LaTeX/reconstruction_pipeline_V4.PDF}
% \caption{\textit{3D Reconstruction and Seed Matching System}}
% \label{fig:reconstruction_pipeline}
% \end{figure*}

\subsection{Reconstruction and Seed Matching} \label{construct}
The main components for the reconstruction and matching pipeline are shown in Fig. \ref{reconstruction_pipeline}: at each time step, a stereo image pair is used to generate a 3D point cloud and predict bounding boxes as described in \ref{slam_front_end}. To mitigate the effect of noise in the robot kinematics, ICP is used to refine the 3D cloud by finding the relative transformation between the current and previous frame. We compare running ICP on the full point clouds vs ICP only on the projected 3D seeds centers. The final 3D reconstructed cloud is then updated.

\subsection{Preliminary Results}

%Preliminary results of the 3D reconstruction pipeline can be seen in Fig. \ref{fig:icp_results}. Captured stereo images are downsampled to 3cm spacing and then passed to the 3D reconstruction pipeline as described in Section \ref{construct}, forming the final point cloud. As seen in Fig. \ref{fig:icp_results}, running ICP only on the centers of detected seeds produces results that are less blurred, brighter, and capture more of the seed surface compared to full-cloud ICP, where we register full point clouds. The comparatively greater blurring in full-cloud ICP is a result of mixing between seed surfaces and inter-seed points in the final 3D reconstruction, and happens because ICP is prone to run into local minima. In contrast, ICP using only seed centers shows comparatively better results since we operate on fewer semantically meaningful points. These results are highlighted in Fig. \ref{fig:cloud_comparison}. 

\begin{figure}[ht!]
\centering
\includegraphics[width=0.75\columnwidth]{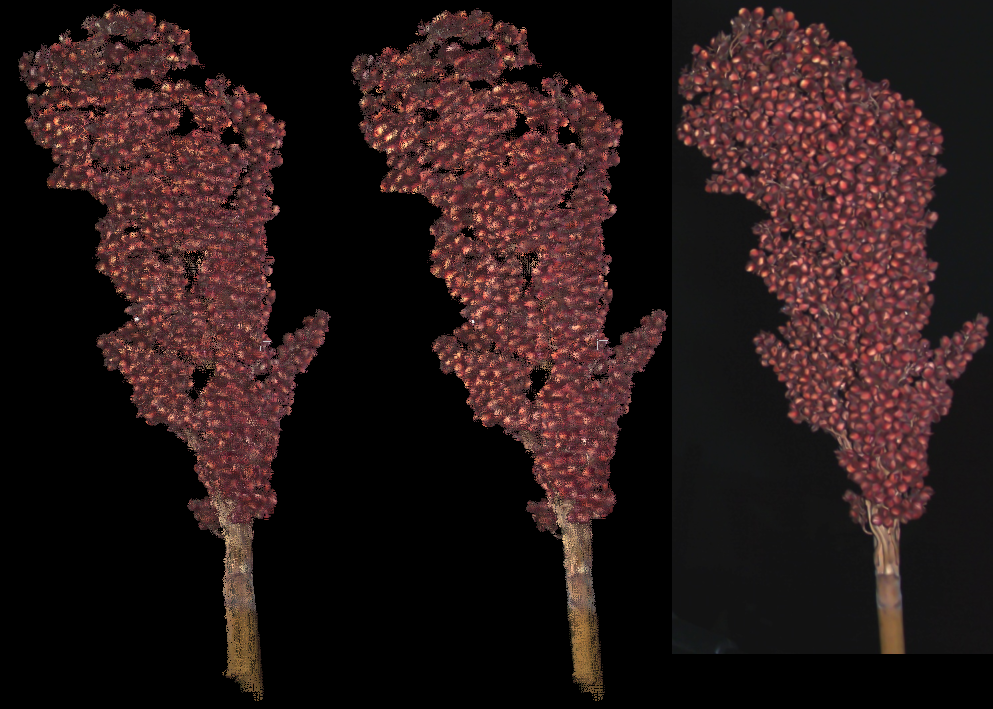} 
\caption{Reconstructions of a sorghum panicle over $90^{\circ}$. Shown are full-cloud ICP (left), ICP on seed centers (middle), and source RGB (right). Enlarged version in appendix.}
\label{fig:icp_results}
\vspace{-10pt}
\end{figure}

\begin{figure}[ht!]
\centering
\includegraphics[width=0.9\columnwidth]{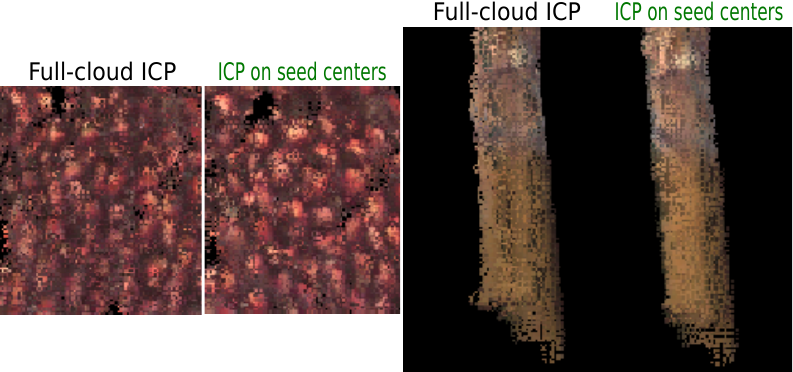} 
\caption{Zoomed view of Fig. \ref{fig:icp_results}.
We see that there is better stem alignment and the reconstructed surface appears brighter (less blurred)  when using ICP on seed centers.}
\label{fig:cloud_comparison}
\vspace{-10pt}
\end{figure}

Captured stereo images using the robotic arm are downsampled to 3cm spacing and then passed to the 3D reconstruction pipeline as described in \ref{construct}, forming the final point cloud. Preliminary 3D reconstruction results can be seen in Fig. \ref{fig:icp_results}. We note that running ICP only on the centers of detected seeds produces results that are less blurred, brighter, and capture more of the seed surface compared to full-cloud ICP, where we register full point clouds. The comparatively greater blurring in full-cloud ICP is a result of mixing between seed surfaces and inter-seed points in the final 3D reconstruction, and occurs because ICP is prone to run into local minima. In contrast, ICP using only seed centers shows comparatively better results since we operate on fewer semantically meaningful points (seed centers). These results are highlighted in Fig. \ref{fig:cloud_comparison}. Note that there are black areas in the 3D reconstruction not present in the RGB image. This is a result of invalid disparity values calculated by SGBM \cite{sgbm}. An interesting direction is to explore is deep-learning depth generation networks such as FCRN Depth Prediction \cite{deeper-depth-pred}.

\section{Future Work}

The SLAM results and initial 3D reconstructions show promise, but there is still much to do to fully realize the benefits of semantic reasoning. Going forward, both SLAM and reconstruction efforts could be improved by considering additional relevant information in the feature matching step. For example, integrating wind speed into the optimization formulation and incorporating learned deep visual features such as \cite{wang2020learning} for a more robust data association process on semantic landmarks. In future work, we also plan to explore high-resolution reconstructions with a larger dataset, containing more varied panicles. An important issue which will need to be addressed is how to assess both the 3D reconstruction and seed match steps in a scaleable manner without ground truth. Eventually, we envision that a better understanding of the scene semantics will allow for motion planning of robotic arms over semantic occupancy maps which could lead to safer and efficient manipulation in various agricultural settings. An interesting question is how to combine scene understanding with reasoning about environment dynamics to create robotic systems that can confidently perform complex manipulation tasks.

% We plan on investigating integrating visual feature descriptors into point matching costs, exploring outlier-robust matching options, and limiting the points considered by ICP to high-confidence and/or geometrically extreme features for better visiblity under rotation.
% For match verification, a simple heuristic that new seeds should mostly appear over the rotating horizon could be used in assessment.

\section*{Acknowledgments}
This work was supported by DOE APRAE TERRA 3P and USDA NIFA CPS 2020-67021-31531. Thanks to Clemson University's Pee Dee Research Center for assisting with data capture. Thanks to CMU PhD students John Kim and Mark Lee for designing and building the camera mount.

\bibliography{main}

\begin{thebibliography}{31}
\providecommand{\natexlab}[1]{#1}

\bibitem[{Baweja et~al.(2018)Baweja, Parhar, Mirbod, and Nuske}]{stalknet}
Baweja, H.~S.; Parhar, T.; Mirbod, O.; and Nuske, S. 2018.
\newblock StalkNet: A Deep Learning Pipeline for High-Throughput Measurement of Plant Stalk Count and Stalk Width.
\newblock In \emph{Field and Service Robotics}, 271--284. Springer.

\bibitem[{Bochkovskiy, Wang, and Liao(2020)}]{yolov4}
Bochkovskiy, A.; Wang, C.; and Liao, H.~M. 2020.
\newblock YOLOv4: Optimal Speed and Accuracy of Object Detection.
\newblock \emph{CoRR}, abs/2004.10934.

\bibitem[{Burkard, Dell'Amico, and Martello(2012)}]{assignment}
Burkard, R.; Dell'Amico, M.; and Martello, S. 2012.
\newblock \emph{Assignment Problems: revised reprint}.
\newblock SIAM.

\bibitem[{Campos et~al.(2021)Campos, Elvira, Rodr{\'\i}guez, Montiel, and Tard{\'o}s}]{campos2021orb}
Campos, C.; Elvira, R.; Rodr{\'\i}guez, J. J.~G.; Montiel, J.~M.; and Tard{\'o}s, J.~D. 2021.
\newblock ORB-SLAM3: An Accurate Open-Source Library for Visual, Visual--Inertial, and Multimap SLAM.
\newblock \emph{IEEE Transactions on Robotics}.

\bibitem[{Choudhary et~al.(2014)Choudhary, Trevor, Christensen, and Dellaert}]{choudhary2014slam}
Choudhary, S.; Trevor, A.~J.; Christensen, H.~I.; and Dellaert, F. 2014.
\newblock SLAM with object discovery, modeling and mapping.
\newblock In \emph{2014 IEEE/RSJ International Conference on Intelligent Robots and Systems}, 1018--1025. IEEE.

\bibitem[{Dong, Roy, and Isler(2020)}]{dong2020semantic}
Dong, W.; Roy, P.; and Isler, V. 2020.
\newblock Semantic mapping for orchard environments by merging two-sides reconstructions of tree rows.
\newblock \emph{Journal of Field Robotics}, 37(1): 97--121.

\bibitem[{Engel, Sch{\"o}ps, and Cremers(2014)}]{lsd-slam}
Engel, J.; Sch{\"o}ps, T.; and Cremers, D. 2014.
\newblock LSD-SLAM: Large-scale direct monocular SLAM.
\newblock In \emph{European conference on computer vision}, 834--849. Springer.

\bibitem[{He et~al.(2017)He, Gkioxari, Dollar, and Girshick}]{Mask-rcnn}
He, K.; Gkioxari, G.; Dollar, P.; and Girshick, R. 2017.
\newblock Mask R-CNN.
\newblock In \emph{2017 IEEE International Conference on Computer Vision (ICCV)}, 2980--2988.

\bibitem[{Hirschmuller(2008)}]{sgbm}
Hirschmuller, H. 2008.
\newblock Stereo Processing by Semiglobal Matching and Mutual Information.
\newblock \emph{IEEE Transactions on Pattern Analysis and Machine Intelligence}, 30(2): 328--341.

\bibitem[{Isola et~al.(2017)Isola, Zhu, Zhou, and Efros}]{pix2pix}
Isola, P.; Zhu, J.-Y.; Zhou, T.; and Efros, A.~A. 2017.
\newblock Image-to-Image Translation with Conditional Adversarial Networks.
\newblock In \emph{Proceedings of the IEEE conference on computer vision and pattern recognition}, 1125--1134.

\bibitem[{Laina et~al.(2016)Laina, Rupprecht, Belagiannis, Tombari, and Navab}]{deeper-depth-pred}
Laina, I.; Rupprecht, C.; Belagiannis, V.; Tombari, F.; and Navab, N. 2016.
\newblock Deeper Depth Prediction with Fully Convolutional Residual Networks.
\newblock \emph{CoRR}, abs/1606.00373.

\bibitem[{Liu et~al.(2016)Liu, Anguelov, Erhan, Szegedy, Reed, Fu, and Berg}]{ssd}
Liu, W.; Anguelov, D.; Erhan, D.; Szegedy, C.; Reed, S.; Fu, C.-Y.; and Berg, A.~C. 2016.
\newblock SSD: Single Shot MultiBox Detector.
\newblock In \emph{European conference on computer vision}, 21--37. Springer.

\bibitem[{Liu et~al.(2018)Liu, Chen, Liu, Shivakumar, Das, Taylor, Underwood, and Kumar}]{monofruit}
Liu, X.; Chen, S.~W.; Liu, C.; Shivakumar, S.~S.; Das, J.; Taylor, C.~J.; Underwood, J.~P.; and Kumar, V. 2018.
\newblock Monocular Camera Based Fruit Counting and Mapping with Semantic Data Association.
\newblock \emph{CoRR}, abs/1811.01417.

\bibitem[{Lowe(2004)}]{sift}
Lowe, D.~G. 2004.
\newblock Distinctive Image Features from Scale-Invariant Keypoints.
\newblock \emph{Int. J. Comput. Vision}, 60(2): 91--110.

\bibitem[{McCormac et~al.(2018)McCormac, Clark, Bloesch, Davison, and Leutenegger}]{fusion++}
McCormac, J.; Clark, R.; Bloesch, M.; Davison, A.; and Leutenegger, S. 2018.
\newblock Fusion++: Volumetric Object-Level SLAM.
\newblock In \emph{2018 international conference on 3D vision (3DV)}, 32--41. IEEE.

\bibitem[{Mur-Artal and Tard\'os(2017)}]{orbslam2}
Mur-Artal, R.; and Tard\'os, J.~D. 2017.
\newblock {ORB-SLAM2}: an Open-Source {SLAM} System for Monocular, Stereo and {RGB-D} Cameras.
\newblock \emph{IEEE Transactions on Robotics}, 33(5): 1255--1262.

\bibitem[{Nellithimaru and Kantor(2019)}]{rols}
Nellithimaru, A.~K.; and Kantor, G.~A. 2019.
\newblock ROLS: Robust Object-level SLAM for Grape Counting.
\newblock In \emph{Proceedings of the IEEE Conference on Computer Vision and Pattern Recognition Workshops}.

\bibitem[{Nicholson, Milford, and S{\"u}nderhauf(2018)}]{quadric}
Nicholson, L.; Milford, M.; and S{\"u}nderhauf, N. 2018.
\newblock Quadricslam: Dual Quadrics from Object Detections as Landmarks in Object-Oriented SLAM.
\newblock \emph{IEEE Robotics and Automation Letters}, 4(1): 1--8.

\bibitem[{Ok et~al.(2019)Ok, Liu, Frey, How, and Roy}]{ok2019robust}
Ok, K.; Liu, K.; Frey, K.; How, J.~P.; and Roy, N. 2019.
\newblock Robust object-based slam for high-speed autonomous navigation.
\newblock In \emph{2019 International Conference on Robotics and Automation (ICRA)}, 669--675. IEEE.

\bibitem[{Parhar et~al.(2018)Parhar, Baweja, Jenkins, and Kantor}]{parharcgan}
Parhar, T.; Baweja, H.; Jenkins, M.; and Kantor, G. 2018.
\newblock A Deep Learning-Based Stalk Grasping Pipeline.
\newblock In \emph{2018 IEEE International Conference on Robotics and Automation (ICRA)}, 1--5. IEEE.

\bibitem[{Qadri(2021)}]{Qadri-2021-128985}
Qadri, M. 2021.
\newblock \emph{Robotic Vision for 3D Modeling and Sizing in Agriculture}.
\newblock Master's thesis, Carnegie Mellon University, Pittsburgh, PA.

\bibitem[{Qadri and Kantor(2021)}]{qadri2021semantic}
Qadri, M.; and Kantor, G. 2021.
\newblock Semantic Feature Matching for Robust Mapping in Agriculture.
\newblock \emph{arXiv preprint arXiv:2107.04178}.

\bibitem[{Ren et~al.(2015)Ren, He, Girshick, and Sun}]{Faster-RCNN}
Ren, S.; He, K.; Girshick, R.; and Sun, J. 2015.
\newblock Faster r-cnn: Towards real-time object detection with region proposal networks.
\newblock In \emph{Advances in neural information processing systems}, 91--99.

\bibitem[{Rublee et~al.(2011)Rublee, Rabaud, Konolige, and Bradski}]{orb}
Rublee, E.; Rabaud, V.; Konolige, K.; and Bradski, G.~R. 2011.
\newblock ORB: An efficient alternative to SIFT or SURF.
\newblock In Metaxas, D.~N.; Quan, L.; Sanfeliu, A.; and Gool, L.~V., eds., \emph{ICCV}, 2564--2571. IEEE Computer Society.
\newblock ISBN 978-1-4577-1101-5.

\bibitem[{Santos et~al.(2020)Santos, {de Souza}, {dos Santos}, and Avila}]{SANTOS2020105247}
Santos, T.~T.; {de Souza}, L.~L.; {dos Santos}, A.~A.; and Avila, S. 2020.
\newblock Grape detection, segmentation, and tracking using deep neural networks and three-dimensional association.
\newblock \emph{Computers and Electronics in Agriculture}, 170: 105247.

\bibitem[{Sep{\'u}lveda et~al.(2020)Sep{\'u}lveda, Fern{\'a}ndez, Navas, Armada, and Gonz{\'a}lez-De-Santos}]{sepulveda2020robotic}
Sep{\'u}lveda, D.; Fern{\'a}ndez, R.; Navas, E.; Armada, M.; and Gonz{\'a}lez-De-Santos, P. 2020.
\newblock Robotic aubergine harvesting using dual-arm manipulation.
\newblock \emph{IEEE Access}, 8: 121889--121904.

\bibitem[{Silwal et~al.(2021)Silwal, Parhar, Yand{\'u}n, and Kantor}]{abhi_stereo}
Silwal, A.; Parhar, T.; Yand{\'u}n, F.; and Kantor, G.~A. 2021.
\newblock A Robust Illumination-Invariant Camera System for Agricultural Applications.
\newblock \emph{ArXiv}, abs/2101.02190.

\bibitem[{Sodhi et~al.(2018)Sodhi, Sun, P{\'o}czos, and Wettergreen}]{sodhi2018robust}
Sodhi, P.; Sun, H.; P{\'o}czos, B.; and Wettergreen, D. 2018.
\newblock Robust plant phenotyping via model-based optimization.
\newblock In \emph{2018 IEEE/RSJ International Conference on Intelligent Robots and Systems (IROS)}, 7689--7696. IEEE.

\bibitem[{Wang et~al.(2020)Wang, Zhou, Hariharan, and Snavely}]{wang2020learning}
Wang, Q.; Zhou, X.; Hariharan, B.; and Snavely, N. 2020.
\newblock Learning feature descriptors using camera pose supervision.
\newblock In \emph{European Conference on Computer Vision}, 757--774. Springer.

\bibitem[{Zhao et~al.(2017)Zhao, Shi, Qi, Wang, and Jia}]{pspnet}
Zhao, H.; Shi, J.; Qi, X.; Wang, X.; and Jia, J. 2017.
\newblock Pyramid Scene Parsing Network.
\newblock In \emph{Proceedings of the IEEE conference on computer vision and pattern recognition}, 2881--2890.

\bibitem[{Zine-El-Abidine et~al.(2021)Zine-El-Abidine, Dutagaci, Galopin, and Rousseau}]{zine2021assigning}
Zine-El-Abidine, M.; Dutagaci, H.; Galopin, G.; and Rousseau, D. 2021.
\newblock Assigning apples to individual trees in dense orchards using 3D colour point clouds.
\newblock \emph{Biosystems Engineering}, 209: 30--52.

\end{thebibliography}

\onecolumn
\appendix
\section*{Appendix}

\begin{figure*}[h]
\begin{center}
  \includegraphics[width=1\linewidth, height=2in]{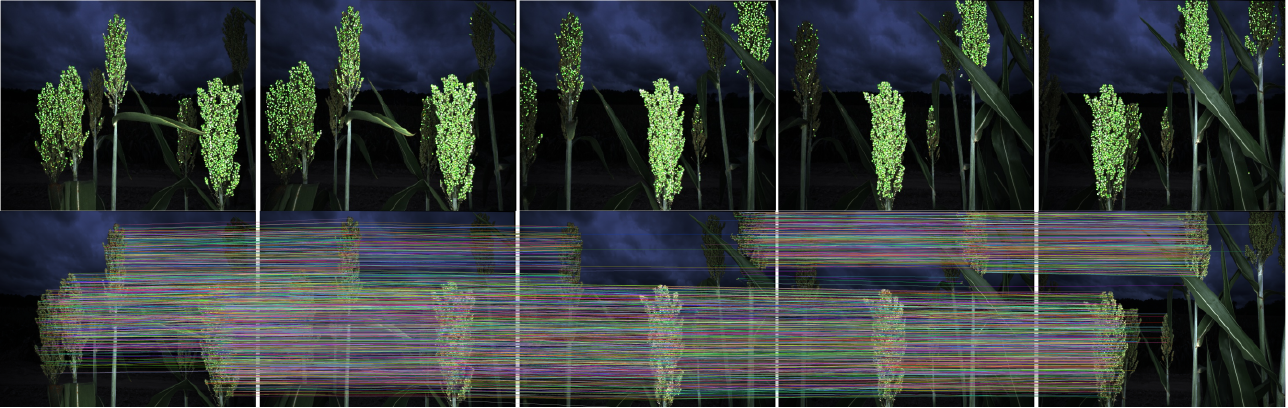}
\end{center}
  \caption{The first row shows the bounding box detections for the SLAM system presented in Section \ref{sectionslam}, one bounding box per sorghum seed. The second row shows the output of the proposed data association pipeline for five consecutive images.}
\label{fig:series}
\end{figure*}

\begin{figure*}[h]
\begin{center}
  \includegraphics[width=0.75\linewidth]{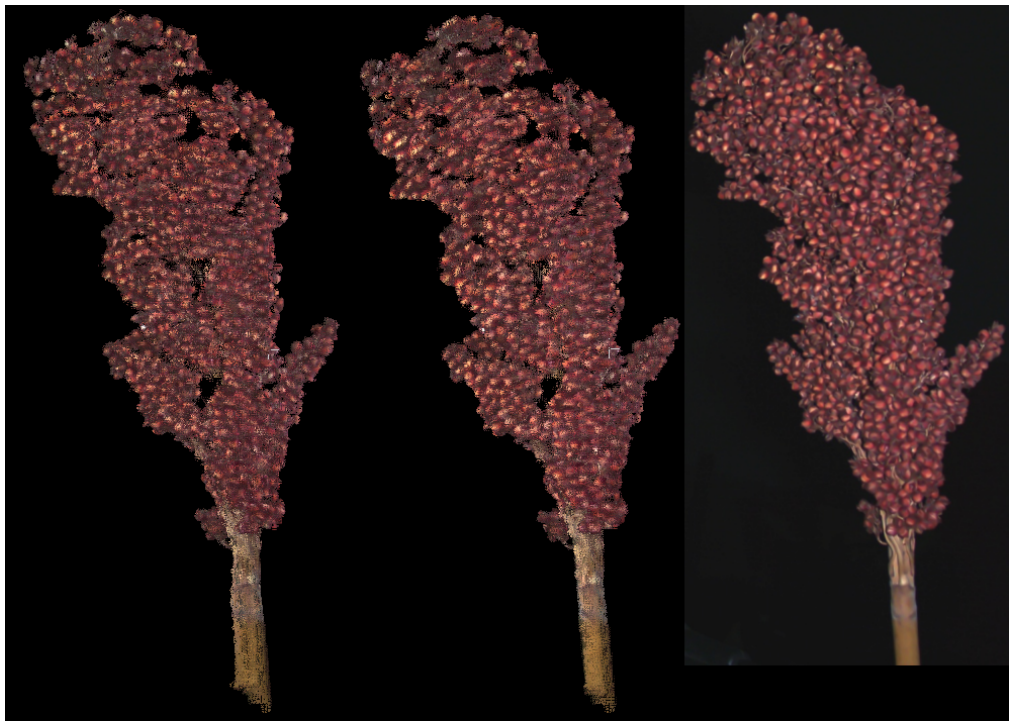}
\end{center}
  \caption{Full-size version Fig. \ref{fig:icp_results}. Shown are full-cloud ICP (left), ICP on seed centers (middle), and source RGB image (right).}
\label{fig:appendix_full_page}
\end{figure*}

% \begin{figure*}[h]
% \includegraphics[width=0.7\linewidth]{LaTeX/cloud_comp_three_img.png} 
% \caption{Full-page version Fig. \ref{fig:icp_results}. Shown are full-cloud ICP (left), ICP on seed centers (middle), and source RGB image (right).}
% \label{fig:appendix_full_page}
% \end{figure*}

\end{document}